\documentclass[runningheads]{llncs}

\usepackage[T1]{fontenc}
\usepackage{lmodern}        
\usepackage{graphicx}
\usepackage{amsmath}
\usepackage{amssymb}
\usepackage{mathtools}
\usepackage{bm}             
\usepackage{multirow}
\usepackage{booktabs}
\usepackage{subcaption}
\usepackage{float}
\usepackage{enumitem}
\usepackage{xr-hyper}
\usepackage{marvosym}

\usepackage{cite}

\usepackage{color}
\usepackage{placeins}
\usepackage{placeins}
\usepackage{hyperref}
\hypersetup{hidelinks}

\usepackage[capitalize,noabbrev]{cleveref}

\def\method{CSGO}

\begin{document}

\title{Continuous Spiking Graph ODE Networks}
\titlerunning{Continuous Spiking Graph ODE Networks}

\author{
Shiqi Fan \and
Zeqing Zhang \and
Nan Yin \and
Tong Li \and
Hongyi Nie \and
Die Hu \and
Wen Hua
}
\institute{}
\maketitle

\begin{abstract}
Spiking Graph Networks (SGNs) represent a biologically inspired paradigm that combines the energy-efficient properties of Spiking Neural Networks (SNNs) with the relational modeling capabilities of Graph Neural Networks (GNNs). Despite their potential, existing SGNs are largely confined to static settings with real-valued or binary inputs, lacking the capacity to model continuous and dynamic graph evolution. Moreover, the discrete nature of spike-based representations introduces significant information loss, limiting predictive performance. In this work, we identify a critical gap in the literature: the absence of a unified, theoretically grounded framework for energy-efficient dynamic graph learning that leverages spiking dynamics. To this end, we propose Continuous Spiking Graph ODE Networks (CSGO), a novel framework that embeds SGNs within the continuous-time dynamics of Graph Neural Ordinary Differential Equations (Graph ODEs). CSGO bridges the temporal spike-latency in SNNs with the continuous evolution of graph structures via ODEs, enabling both dynamic modeling and low power consumption. To address the intrinsic information loss in spiking representations, we further derive a second-order spiking mechanism and formulate the second-order CSGO, enhancing representational capacity. We provide theoretical analysis demonstrating the gradient stability of our framework, and validate its effectiveness through extensive experiments on benchmark datasets. CSGO achieves state-of-the-art performance while significantly reducing energy consumption, offering a principled and scalable solution for neuromorphic dynamic graph learning.

\end{abstract}

\section{Introduction}
\begin{figure}
    \centering
    \includegraphics[width=0.7\linewidth]{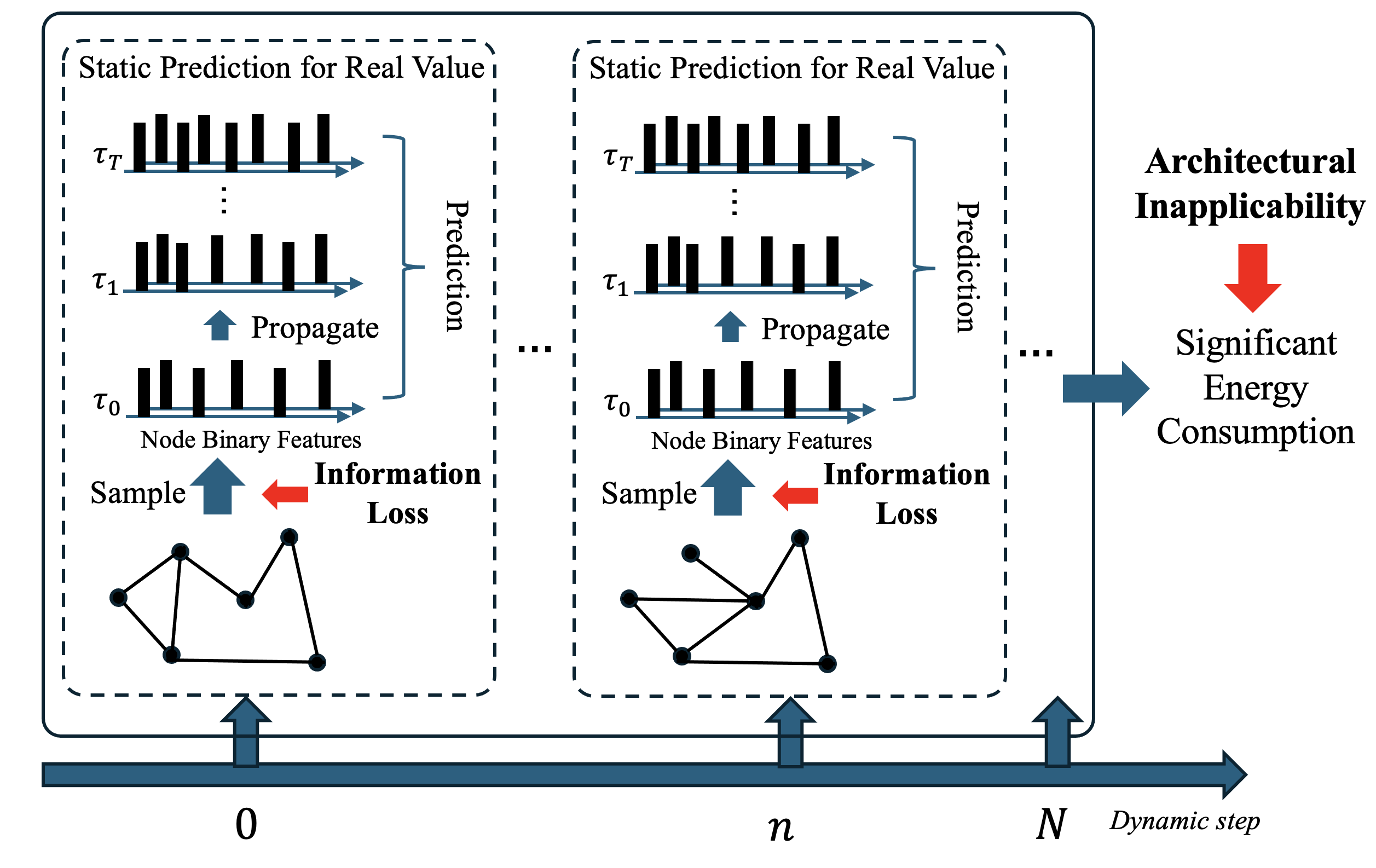}
    \caption{Illustration of information loss and architecture inapplicability. Information loss arises from the sampling of real-value inputs, while the architecture inapplicability refers to the unsuitability of dynamic real-valued SGNs.}
    \label{fig:challenges}
\end{figure}

Spiking Graph Networks (SGNs)~\cite{zhu2022spiking,xu2021exploiting} combine the structural modelling capability of GNNs with the event-driven, low-power computation of Spiking Neural Networks (SNNs), and have proved effective in object recognition, real-time data analysis, and static graph classification~\cite{li2023scaling,zhu2022spiking,xu2021exploiting}.

However, existing SGNs are designed for \emph{static} graphs and rely on Bernoulli sampling of real-valued inputs into binary spikes. Re-sampling at every step on dynamic inputs is expensive, undermining the SNN energy advantage, and the binarization itself discards fine-grained temporal and semantic information~\cite{yan2021near}.
As illustrated in Figure~\ref{fig:challenges}, this leads to two obstacles: (i) \emph{architectural inapplicability} to continuously evolving real-valued inputs, and (ii) \emph{semantic information loss} from discrete spike encoding.

Graph Neural Ordinary Differential Equations (Graph ODEs) offer a continuous-time tool for dynamic graph representation, while high-order neural mechanisms~\cite{luo2023hope} can recover information lost in low-order encodings. Combining the two is, however, non-trivial: SGNs evolve over spike latency whereas Graph ODEs evolve over physical time; spike encoding loses semantic detail; and continuous-depth architectures are prone to vanishing or exploding gradients.

We propose Continuous Spiking Graph ODE Networks (\textsc{CSGO}) to address all three issues. We first formulate SGNs as a component of continuous-time ODEs and obtain a first-order variant \textsc{CSGO}-1st, then design a second-order spike encoding mechanism and embed it into a second-order Graph ODE, yielding \textsc{CSGO}-2nd. We further show that gradients remain bounded throughout training. Extensive experiments on node- and graph-classification benchmarks confirm that \textsc{CSGO} delivers state-of-the-art accuracy with substantially lower energy cost.

In summary, our contributions are:
\begin{itemize}[itemsep=2pt,topsep=0pt,parsep=0pt]
\item We identify the limitations of existing SGNs on dynamic graph tasks and define a new setting requiring both continuous temporal modelling and low-power spike-based computation.
\item We propose \method{}, a unified SGN--Graph ODE framework, with theoretical guarantees that it avoids vanishing and exploding gradients.
\item Comprehensive experiments show that \method{} consistently outperforms state-of-the-art GNN, SNN, and Graph ODE baselines on both accuracy and energy.
\end{itemize}

\section{Related Work}

\paragraph{Graph Ordinary Differential Equations.}
Dynamic GNNs that initialize node states with discrete-timestamp embeddings~\cite{kipf2018neural,chen2018neural} typically require all nodes to be present at every timestamp~\cite{huang2021coupled}, which is impractical in real-world settings.
Graph ODEs avoid this by treating evolution as a continuous process and tolerate missing observations~\cite{chen2018neural}; recent variants~\cite{poli2019graph} couple GNN-encoded states with neural ODEs over both nodes and edges, while high-order extensions~\cite{luo2023hope} better capture long-range temporal dependencies.
We embed energy-efficient SNNs into the Graph ODE framework, retaining the low-power advantage of spiking computation while leveraging the dynamic modelling capacity of ODEs.

\paragraph{Spiking Graph Networks.}
Spiking Graph Networks (SGNs)~\cite{zhu2022spiking,xu2021exploiting} reduce the energy cost of graph learning by replacing real-valued activations with reactive spiking neurons, but are typically restricted to static graphs with real-valued or binary inputs.
Existing attempts to apply static SNNs to dynamic graphs~\cite{li2023scaling,yin2024dynamic} re-encode the spike representation at every step, which negates the energy benefit of SNNs.
Our framework instead unifies SGNs with Graph ODEs, capturing dynamic evolution while preserving low-power operation.

\section{Preliminaries}

\subsection{Problem Definition}

We study the problem of energy-efficient dynamic graph classification. Given a temporal graph sequence $\mathcal{G} = \{G^{1}, G^{2}, \dots, G^{T}\}$, where each graph snapshot $G^{t} = (V^{t}, E^{t})$ contains a node feature matrix $\mathbf{X}^{t} \in \mathbb{R}^{|V^{t}| \times d}$, the goal is to predict a target label $\hat{y} \in \mathcal{Y}$ by learning a function $f_\theta$ such that $\hat{y} = f_\theta(\mathcal{G})$. To achieve this under neuromorphic energy constraints, we encode each real-valued feature $\mathbf{X}^{t}_{i,j}$ into a binary spike train $\mathbf{S}^{t}_{i,j} \in \{0,1\}^{N}$ across N latency steps using stochastic Bernoulli sampling:
$\mathbf{S}^{t}_{i,j,\tau} \sim \text{Bernoulli}(\mathbf{X}^{t}_{i,j}), \quad \tau = 1, \dots, N$.
This yields a spiking input tensor $\mathbf{S}^{t} \in \{0,1\}^{|V^{t}| \times d \times N}$ that reflects the temporal firing activity of each node. These spike trains are then used as input to a spiking graph encoder, which integrates structural message passing over $G^{t}$ with temporal spike dynamics to compute node embeddings $\mathbf{H}^{t}$. A temporal evolution model—e.g., based on neural ODEs—then fuses the sequence $\{\mathbf{H}^{1}, \dots, \mathbf{H}^{T}\}$ into a unified representation $\mathbf{z}$, from which a classifier predicts the final output: $\hat{y} = g_\theta(\mathbf{z})$. This pipeline presents key challenges in accurately capturing graph-temporal dependencies while preserving energy efficiency and representation stability, which we address through our proposed continuous spiking graph ODE framework.

\subsection{Continuous Graph Neural Networks}
\label{sec:cgnn}
Let $G=(\mathcal{V},\mathcal{E})$ with feature matrix $\mathbf{X}\in\mathbb{R}^{|\mathcal{V}|\times d}$ and adjacency $\mathbf{A}$.
The first ODE-based CGNN~\cite{xhonneux2020continuous} extends the simple GNN $\mathbf{H}_{n+1}=\mathbf{A}\mathbf{H}_n+\mathbf{H}_0$ to continuous time, giving
\begin{equation}
\label{first_graph_ode}
    \tfrac{d\mathbf{H}(t)}{dt}=\ln\mathbf{A}\,\mathbf{H}(t)+\mathbf{E},
\end{equation}
where $\mathbf{E}=\varepsilon(\mathbf{X})$ is an encoder output.
GraphCON~\cite{rusch2022graph} introduces a second-order Graph ODE $\mathbf{X}''=\sigma(\mathbf{F}_{\theta}(\mathbf{X},t))-\gamma \mathbf{X}-\alpha \mathbf{X}'$ with a learnable coupling function $\mathbf{F}_\theta$, and solves it with the IMEX (implicit-explicit) symplectic-Euler scheme:
\begin{equation}
\label{second_graph_ode}
\begin{aligned}
    \mathbf{Y}^n&=\mathbf{Y}^{n-1}+\Delta t[\sigma(\mathbf{F}_\theta(\mathbf{X}^{n-1},t^{n-1}))-\gamma\mathbf{X}^{n-1}-\alpha\mathbf{Y}^{n-1}],\\
    \mathbf{X}^n&=\mathbf{X}^{n-1}+\Delta t\,\mathbf{Y}^n,\quad n=1,\cdots,T,
\end{aligned}
\end{equation}
with fixed time-step $\Delta t>0$ and hidden features $\mathbf{Y}^n,\mathbf{X}^n$ at $t^n=n\Delta t$.

\subsection{Spiking Neural Networks}

\textbf{First-order SNNs.}
SNNs convert real-valued inputs into binary spikes over time; each neuron accumulates input spikes into a membrane potential and fires when the threshold $V_{th}$ is exceeded:
\begin{equation}
\label{first-order-snn}
    u_{\tau+1,i} = \lambda (u_{\tau,i} - V_{th} s_{\tau,i}) + \textstyle\sum_{j} w_{ij} s_{\tau,j} + b ,\qquad
    s_{\tau+1,i} = \mathbb{H}(u_{\tau+1,i} - V_{th}),
\end{equation}
where $\mathbb{H}(\cdot)$ is the non-differentiable Heaviside function,
$s_{\tau,i}\in\{0,1\}$ is the binary spike, and $\lambda$ is a leak constant.

\textbf{Second-order SNNs.}
First-order models assume an instantaneous current response, ignoring the gradual neurotransmitter dynamics observed in real synapses.
The synaptic-conductance-based LIF model~\cite{eshraghian2023training} introduces a second-order recurrence with separate synaptic and membrane time constants:
\begin{equation}
\label{second-order}
    I_{\tau+1}=\alpha I_{\tau}+W X_{\tau+1},\quad
    u_{\tau+1,i}=\beta u_{\tau,i}+I_{\tau+1,i}-V_{th}s_{\tau,i},\quad
    s_{\tau+1,i}=\mathbb{H}(u_{\tau+1,i}-V_{th}),
\end{equation}
where
$\alpha=\exp(-\Delta\tau/\tau_{syn})$ and
$\beta=\exp(-\Delta\tau/\tau_{mem})$
control the synaptic-current and membrane-potential decay, respectively.

\section{Methodology}
\begin{figure}
  \centering
  \includegraphics[scale=0.4]{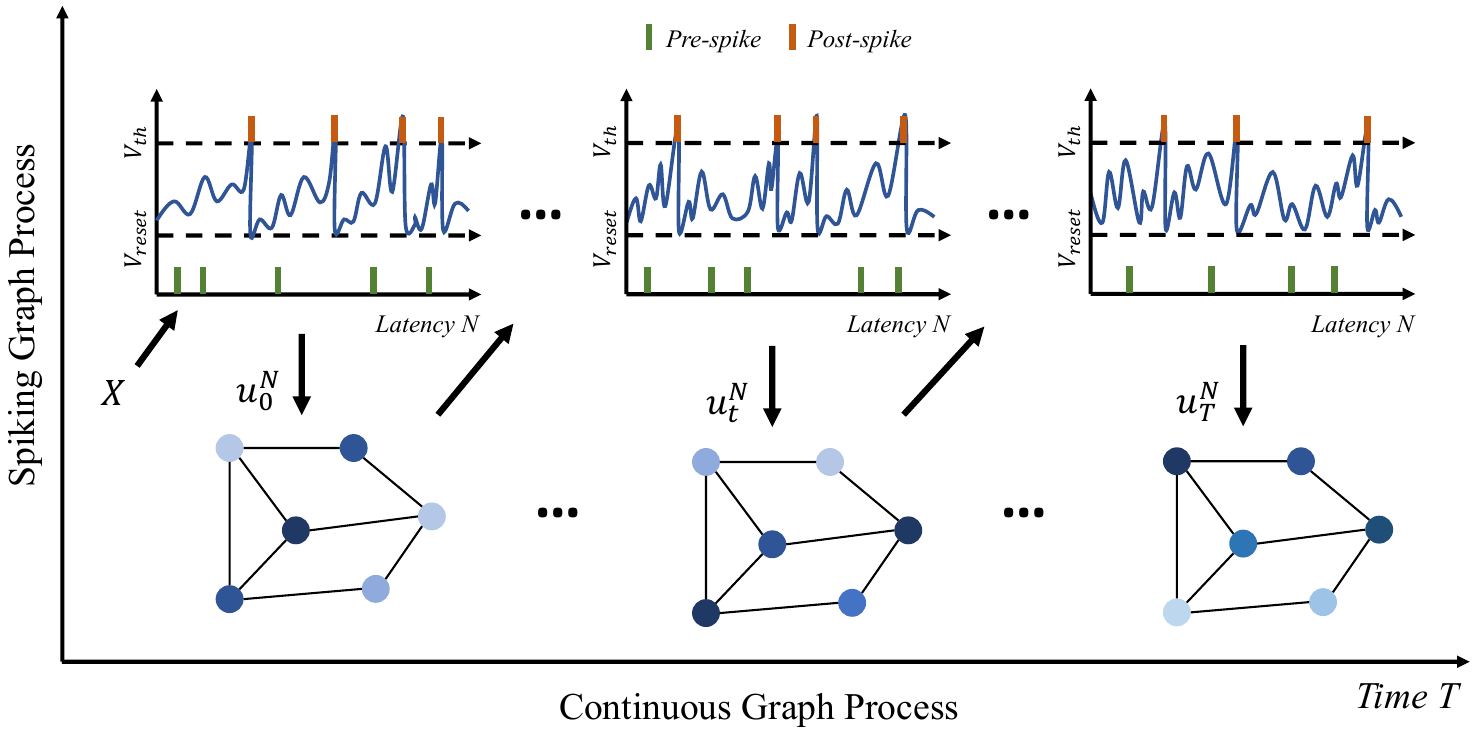}
  \caption{Overview of the proposed \method{}. The proposed \method{} takes a graph with node features as input, which are initially encoded using the SNNs (first-order or second-order). Subsequently, a first-order or second-order continuous graph process is employed to evolve the representation over different time step. Finally, the representation is projected for downstream tasks.}
  \label{framework}
\end{figure}

In this part, we present the proposed \method{} for modeling the continuous spiking graph neural networks.
\method{} combines SNNs with CGNNs in a framework of PDE, which preserves the advantage of SNNs and CGNNs simultaneously. To mitigate the problem of information loss attributed to SNNs, we involve the derivation of second-order spike representation and differentiation for second-order SNNs, and then introduce the high-order PDE, referred to as \method{}-2nd. Finally, we present a theoretical proof to ensure that our proposed \method{} effectively mitigates the challenges associated with gradient exploding and vanishing. The framework of \method{} is shown in Figure~\ref{framework}.



\subsection{First-order \method{}}
Specifically, SNNs propagate information within the time latency $\tau$, and the CGNNs evaluate the feature evolution along with time $t$. We propose the first-order \method{} to incorporate SNNs with CGNNs in the form of PDE with their respective timelines. Intuitively, information is interactively propagated in dual time dimensions by both the SNNs and the CGNNs, which is formulated as: 
\begin{proposition}
\label{proposition_first}
Define the first-order SNNs as
$\frac{d u_t^\tau}{d \tau}=g(u_t^\tau,\tau)$,
and first-order CGNNs as
$\frac{d u_t^\tau}{d t}=f(u_t^\tau,t)$.
Let
$A_y=\int_0^N g(u_y^x,x)\,dx$.
Then the first-order \method{} can be formulated as
\begin{equation}
\label{f_g}
u_T^N
=
2\int_0^{T-1} f(A_y)\,dy
+
\int_{T-1}^{T} f(A_y)\,dy .
\end{equation}
where $N$ is the total latency of SNNs, $T$ is the number of CGNN time steps,
and $u_y^x$ denotes the neuron membrane potential at latency
$x\in[0,N]$ and time step $y\in[0,T]$.
Details of the derivation are provided in the Appendix.
\end{proposition}

To provide a comprehensive understanding of Proposition~\ref{proposition_first}, we approach it from two distinct perspectives. In Eq.~\eqref{f_g}, within the temporal dimension where $0\leq t \leq T$, the SNNs initially compute spike representations at each time step $t$. These representations are then employed in the evolution of the CGNNs process. Then, the CGNNs model the evolution of node embeddings along a time series $t$ and subsequently integrates them across the latency dimension.

In our implementation of the first-order \method{}, we employ Eq.~\eqref{f_g} by initially calculating spike representations at each time step, followed by modeling the evolution of node embeddings. As described in~\cite{meng2022training}, the first-order spike representation at time step $t$ is denoted as: $\mathbf{H}(t)=\frac{\sum_{\tau=1}^N\lambda^{N-\tau}s_\tau^t}{\sum_{\tau=1}^N\lambda^{N-\tau}}$. By combining Eq.~\eqref{first_graph_ode}, we have:
\begin{equation}
    \frac{d\mathbf{H}(t)}{dt}=ln\mathbf{AH}(t)+\frac{\sum_{\tau=1}^N\lambda^{N-\tau}s_\tau^0}{\sum_{\tau=1}^N\lambda^{N-\tau}},
\end{equation}
where $s_\tau^t$ is the binary spike on latency $\tau$ and time step $t$, and $\lambda=exp(-\frac{\Delta\tau}{\kappa})$ with $\Delta\tau \ll \kappa$, $\kappa$ is the time constant. 

\subsection{Second-order Spiking Neural Networks}
The proposed first-order \method{} effectively addresses the challenge of combining SNNs with CGNNs to achieve energy-efficient modeling for continuous graph learning.
However, the first-order SNNs typically suffer from the information loss issue. Motivated by recent advancements in high-order structure modeling~\cite{luo2023hope}, which addresses high-order correlations of nodes, we introduce the second-order \method{} to tackle the aforementioned issue. In this part, we begin by deriving the second-order spike representation and investigating the backpropagation of second-order SNNs.


\textbf{Second-order SNNs Forward Propagation.}
We set the forward propagation layer of SNNs to $L$. According to Eq.~\ref{second-order}, the spiking dynamics of layer $i$ can be written as
$u^i(n)=\beta^i u^i(n-1)+(1-\beta^i)\frac{V_{th}^{i-1}}{\Delta\tau}(\alpha^i I^{i-1}(n-1)+\mathbf{W}^is^{i-1}(n))-V^i_{th}s^i(n)$,
where $i=1,\cdots,L$, $n=1,\cdots,N$ (latency steps),
$\alpha^i=\exp(-\Delta\tau/\tau_{syn}^i)$, $\beta^i=\exp(-\Delta\tau/\tau_{mem}^i)$ and $0<\Delta\tau\ll\{\tau_{syn}^i,\tau_{mem}^i\}$.

\textbf{Second-order Spike Representation.}
Following the first-order spike representation framework~\cite{meng2022training}, we
define the weighted average input current as $\hat{I}(N)=\frac{1}{(\beta-\alpha)^2}\frac{\sum_{n=0}^{N-1}(\beta^{N-n}-\alpha^{N-n})I_{in}(n)}{\sum_{n=0}^{N-1}(\beta^{N-n}-\alpha^{N-n})}$ and the scaled weighted firing rate as $\hat{a}(N)=\frac{1}{\beta^2}\frac{V_{th}\sum_{n=0}^{N-1}\beta^{N-n}s(n)}{\sum_{n=0}^{N-1}(\beta^{N-n}-\alpha^{N-n})\Delta\tau}$.
Unrolling Eq.~\eqref{second-order} and applying the limits $\Delta\tau\to 0$, $\hat{a}(N)\in[0,V_{th}/\Delta\tau]$ together with the small-$V_{th}$ assumption (full derivation in Appendix), we obtain the closed-form spike representation:
\begin{equation}
\label{second_order_representation}
    \lim\limits_{N \to \infty}\hat{a}(N)\approx \mathrm{clamp} \left(\tfrac{\tau_{syn}\tau_{mem}}{\tau_{mem}-\tau_{syn}}\hat{I}(N),0,\tfrac{V_{th}}{\Delta \tau} \right),
\end{equation}
where $\mathrm{clamp}(x,a,b)=\max(a,\min(x,b))$.
This yields Proposition~\ref{proposition_second}, which extends the first-order result of~\cite{meng2022training} to second-order SNNs.
\begin{proposition}
\label{proposition_second}
For $i=1,\ldots,L$, define
\begin{equation}
\label{eq:second_rates}
\begin{aligned}
\hat{a}^0(N)
&=
\frac{
\sum_{n=0}^{N-1}(\beta_i)^{N-n-2}s^0(n)
}{
\sum_{n=0}^{N-1}
\left((\beta_i)^{N-n}-(\alpha_i)^{N-n}\right)\Delta\tau
},\\
\hat{a}^i(N)
&=
\frac{
V_{th}^i\sum_{n=0}^{N-1}(\beta_i)^{N-n-2}s^i(n)
}{
\sum_{n=0}^{N-1}
\left((\beta_i)^{N-n}-(\alpha_i)^{N-n}\right)\Delta\tau
}.
\end{aligned}
\end{equation}
where
\begin{equation}
\label{eq:second_alpha_beta}
\alpha_i=\exp\left(-\frac{\Delta\tau}{\tau_{syn}^i}\right),
\qquad
\beta_i=\exp\left(-\frac{\Delta\tau}{\tau_{mem}^i}\right).
\end{equation}
The differentiable mapping is
\begin{equation}
\label{eq:second_mapping}
\mathbf{z}^i
=
\operatorname{clamp}
\left(
c_i\mathbf{W}^i\mathbf{z}^{i-1},
0,
\frac{V_{th}^i}{\Delta\tau}
\right),
\qquad
c_i=
\frac{\tau_{syn}^i\tau_{mem}^i}
{\tau_{mem}^i-\tau_{syn}^i}.
\end{equation}
If $\lim_{N\to\infty}\hat{a}^i(N)=\mathbf{z}^i$ for $i=0,\ldots,L-1$, then
$\hat{a}^{i+1}(N)\approx\mathbf{z}^{i+1}$ as $N\to\infty$.
\end{proposition}
\textbf{Differentiation on Second-order Spike Representation.}
We define the spike representation operator $r(s)=\frac{1}{\beta^2}\frac{V_{th}\sum_{n=0}^{N-1}\beta^{N-n}s(n)}{\sum_{n=0}^{N-1}(\beta^{N-n}-\alpha^{N-n})\Delta \tau}$ and obtain the layer output $\mathbf{o}^i=r(s^i)\approx\mathrm{clamp}(\mathbf{W}^ir(s^{i-1}),0,V_{th}^i/\Delta\tau)$. The gradient with respect to $\mathbf{W}^i$ then follows from the chain rule:
\begin{equation}
\label{back}
\frac{\partial \mathcal{L}}{\partial \mathbf{W}^i}
=\frac{\partial \mathcal{L}}{\partial \mathbf{o}^{i+1}}\frac{\partial \mathbf{o}^{i+1}}{\partial \mathbf{o}^i}\frac{\partial \mathbf{o}^i}{\partial \mathbf{W}^i},
\end{equation}
which provides a tractable backpropagation rule for second-order SNNs.

\subsection{Second-order \method{}}
Having obtained the second-order spike representation for SNNs, we introduce the second-order \method{}. While obtaining an analytical solution for the second-order \method{} may not be feasible, we can derive a conclusion similar to Proposition~\ref{proposition_first}. The specifics are presented as follows.


\begin{proposition}
\label{proposition_third}
Define the second-order SNNs as
$\frac{d^2u_t^\tau}{d\tau^2}
+\delta\frac{du_t^\tau}{d\tau}
=g(u_t^\tau,\tau)$,
and second-order CGNNs as
$\frac{d^2u_t^\tau}{dt^2}
+\gamma\frac{du_t^\tau}{dt}
=f(u_t^\tau,t)$.
Then the second-order \method{} follows:
\begin{equation}
\begin{aligned}
u_t^\tau
&=
\int_0^T
h\left(
\int_0^N e(u_t^\tau)\,d\tau
\right)dt \\
&=
\int_0^N
e\left(
\int_0^T h(u_t^\tau)\,dt
\right)d\tau,
\nonumber
\end{aligned}
\end{equation}
where
$e(u_t^\tau)=\frac{\partial u_t^\tau}{\partial\tau}$ and
$h(u_t^\tau)=\frac{\partial u_t^\tau}{\partial t}$ satisfy
\[
\frac{\partial e(u_t^\tau)}{\partial\tau}
+\delta e(u_t^\tau)
=g(u_t^\tau),
\qquad
\frac{\partial h(u_t^\tau)}{\partial t}
+\gamma h(u_t^\tau)
=f(u_t^\tau).
\]
Here, $\delta$ and $\gamma$ are the hyperparameters of the second-order
SNNs and CGNNs.
\end{proposition}

The details are derived in Appendix.
Similarly to the first-order \method{}, we implement the second-order \method{} by calculating the spike representation on each time step and then model the node embedding with Eq.~\eqref{second_graph_ode}.
Furthermore, to optimize the second-order \method{}, we analyze the differentiation of the second-order \method{}.
Denote the loss function as $\mathcal{L}=\sum\limits_{i\in\mathcal{V}}\left|\mathbf{X}_i^T-\mathbf{\bar{X}}_i\right|^2$, and $\mathbf{\bar{X}}_i$ is the label of node $i$. With the chain rule, we have: $\frac{\partial \mathcal{L}}{\partial \mathbf{W}^l}=\frac{\partial \mathcal{L}}{\partial \mathbf{o}^N_T} \frac{\partial \mathbf{o}^N_T}{\partial \mathbf{o}^l_T} \frac{\partial \mathbf{o}^l_T}{\partial \mathbf{W}^l}$. From~\cite{rusch2022graph}, we have the conclusion that the traditional GNN model has the problem of gradient exponentially or vanishing, thus, we study the upper bound of the proposed \method{}.

\begin{proposition}
\label{proposition_fourth}
Let $\mathbf{X}^n$ and $\mathbf{Y}^n$ be the node features generated by
Eq.~\eqref{second_graph_ode}, and let $\Delta t \ll 1$.
The gradient of the second-order CGNN parameter $\mathbf{W}_l$ is bounded
as in Eq.~\eqref{bound_gnn}, while the gradient of the second-order SNN
parameter $\mathbf{W}^k$ is bounded as in Eq.~\eqref{bound_snn}:
\begin{equation}
\label{bound_gnn}
\begin{aligned}
\left|
\frac{\partial \mathcal{L}}{\partial \mathbf{W}_l}
\right|
&\leq
\frac{\beta'\hat{D}\Delta t(1+\Gamma T\Delta t)}{v}
\max_{1\leq i\leq v}
\left(
|\mathbf{X}_i^0|+|\mathbf{Y}_i^0|
\right)\\
&\quad+
\frac{\beta'\hat{D}\Delta t(1+\Gamma T\Delta t)}{v}
\left(
\max_{1\leq i\leq v}|\bar{\mathbf{X}}_i|
+\beta\sqrt{T\Delta t}
\right)^2 ,
\end{aligned}
\end{equation}

\begin{equation}
\label{bound_snn}
\left|
\frac{\partial \mathcal{L}}{\partial \mathbf{W}^k}
\right|
\leq
\frac{
(1+T\Gamma\Delta t)
(1+N\Theta\Delta\tau)V_{th}
}{
v\beta^2\Delta\tau
}A .
\end{equation}

where
$\beta=\max_x|\sigma(x)|$,
$\beta'=\max_x|\sigma'(x)|$, and
$\hat{D}=\max_{i,j\in\mathcal{V}}\frac{1}{\sqrt{d_i d_j}}$.
Moreover,
$\Gamma:=6+4\beta'\hat{D}
\max_{1\leq n\leq T}\|\mathbf{W}^n\|_1$
and
$\Theta:=6+4\beta'\hat{D}
\max_{1\leq n\leq N}\|\mathbf{W}^n\|_1$.
Here, $d_i$ is the degree of node $i$,
$\bar{\mathbf{X}}_i$ is the label of node $i$, and
\[
A=
\max_{1\leq i\leq v}|\mathbf{X}_i^T|
+
\max_{1\leq i\leq v}|\bar{\mathbf{X}}_i|.
\]
Eq.~\eqref{bound_gnn} follows directly from~\cite{rusch2022graph},
and the derivation of Eq.~\eqref{bound_snn} is provided in the Appendix.
    \end{proposition}

The upper bound in Proposition \ref{proposition_fourth} demonstrates that the total gradient remains globally bounded, regardless of the number of CGNNs layers $T$ and SNNs layers $N$, as long as $\Delta t \sim T^{-1}$ and $\Delta \tau \sim N^{-1}$. This effectively addresses the issues of exploding and vanishing gradients.

\section{Experiments}

\subsection{Experimental Settings}
\textbf{Datasets.}
For node classification we use three homophilic datasets (Cora~\cite{mccallum2000automating}, Citeseer~\cite{sen2008collective}, Pubmed~\cite{namata2012query}) and three heterophilic ones from WebKB (Texas, Wisconsin, Cornell); homophily levels follow~\cite{pei2020geom} and are listed in Tables~\ref{table_1} and~\ref{table_2}. For graph classification we use the superpixel MNIST dataset~\cite{lecun1998gradient}, where each grayscale image is converted to a 75-vertex graph (vertices = superpixels, edges = spatial adjacency). We follow the standard 55K/5K/10K train/val/test split of~\cite{rusch2022graph}.

\noindent\textbf{Baselines.}
For node classification, we compare CSGO with representative GNNs
such as GCN and GAT~\cite{kipf2017semi,velickovic2017graph},
continuous-depth models including CGNN and GraphCON~\cite{xhonneux2020continuous,rusch2022graph},
heterophily-aware GNNs such as H2GCN~\cite{zhu2020generalizing},
and the spiking baseline SpikingGCN~\cite{zhu2022spiking}.
For graph classification, we additionally include representative architectures
such as GIN~\cite{xu2019powerful}. Baseline results follow the protocol
of GraphCON~\cite{rusch2022graph}.

\begin{table}[t]
\centering
\caption{The test accuracy (in \%) for node classification on homophilic datasets. \textbf{Bold} numbers indicate the best performance, and \underline{underline} indicates the second-best performance.}
\label{table_1}
\begin{tabular*}{\linewidth}{@{\extracolsep{\fill}}lccc}
\toprule
 & Cora & Citeseer & Pubmed \\
\textit{Homophily level} & 0.81 & 0.74 & 0.80 \\
\midrule
GAT-ppr & $81.6\pm0.3$ & $68.5\pm0.2$ & $76.7\pm0.3$ \\
MoNet & $81.3\pm1.3$ & $71.2\pm2.0$ & $78.6\pm2.3$ \\
GraphSage & $79.2\pm7.7$ & $71.6\pm1.9$ & $77.4\pm2.2$ \\
CGNN & $81.4\pm1.6$ & $66.9\pm1.8$ & $66.6\pm4.4$ \\
GDE & $78.7\pm2.2$ & $71.8\pm1.1$ & $73.9\pm3.7$ \\
GCN & $81.5\pm1.3$ & $71.9\pm1.9$ & $77.8\pm2.9$ \\
GAT & $81.8\pm1.3$ & $71.4\pm1.9$ & $78.7\pm2.3$ \\
SGC & $81.5\pm0.4$ & $71.7\pm0.4$ & $79.2\pm0.3$ \\
GRAND & $83.6\pm1.0$ & $73.4\pm0.5$ & $78.8\pm1.7$ \\
GraphCON-GCN & $81.9\pm1.7$ & $72.9\pm2.1$ & $78.8\pm2.6$ \\
GraphCON-GAT & $83.2\pm1.4$ & $73.2\pm1.8$ & $\underline{79.5\pm1.8}$ \\
SpikingGCN & $80.7\pm0.6$ & $72.5\pm0.2$ & $77.6\pm0.5$ \\
\midrule
\method{}-1st & $\underline{83.3\pm2.1}$ & $\underline{73.7\pm2.0}$ & $76.9\pm2.7$ \\
\method{}-2nd & $\mathbf{83.7\pm1.3}$ & $\mathbf{75.2\pm2.0}$ & $\mathbf{79.6\pm2.3}$ \\
\bottomrule
\end{tabular*}
\end{table}

\begin{table}[t]
\centering
\caption{The test accuracy (in \%) for node classification on heterophilic datasets. \textbf{Bold} numbers indicate the best performance, and \underline{underline} indicates the second-best performance.}
\label{table_2}
\begin{tabular*}{\linewidth}{@{\extracolsep{\fill}}lccc}
\toprule
 & Texas & Wisconsin & Cornell \\
\textit{Homophily level} & 0.11 & 0.21 & 0.30 \\
\midrule
GPRGNN & $78.4\pm4.4$ & $82.9\pm4.2$ & $80.3\pm8.1$ \\
H2GCN & $84.9\pm7.2$ & $87.7\pm5.0$ & $82.7\pm5.3$ \\
GCNII & $77.6\pm3.8$ & $80.4\pm3.4$ & $77.9\pm3.8$ \\
Geom-GCN & $66.8\pm2.7$ & $64.5\pm3.7$ & $60.5\pm3.7$ \\
PairNorm & $60.3\pm4.3$ & $48.4\pm6.1$ & $58.9\pm3.2$ \\
GraphSAGE & $82.4\pm6.1$ & $81.2\pm5.6$ & $76.0\pm5.0$ \\
MLP & $80.8\pm4.8$ & $85.3\pm3.3$ & $81.9\pm6.4$ \\
GCN & $55.1\pm5.2$ & $51.8\pm3.1$ & $60.5\pm5.3$ \\
GAT & $52.2\pm6.6$ & $49.4\pm4.1$ & $61.9\pm5.1$ \\
GraphCON-GCN & $\underline{85.4\pm4.2}$ & $\underline{87.8\pm3.3}$ & $\bm{84.3\pm4.8}$ \\
GraphCON-GAT & $82.2\pm4.7$ & $85.7\pm3.6$ & $83.2\pm7.0$ \\
\midrule
\method{}-1st & $81.6\pm6.2$ & $84.9\pm3.2$ & $80.4\pm1.9$ \\
\method{}-2nd & $\bm{87.3\pm4.2}$ & $\bm{88.8\pm2.5}$ & $\underline{83.7\pm2.7}$ \\
\bottomrule
\end{tabular*}
\end{table}

\noindent\textbf{Implementation Details.}
We report mean accuracy over 20 random initializations on 5 splits for homophilic datasets and over 10 fixed splits for heterophilic ones; baseline numbers follow~\cite{rusch2022graph}. We set $\lambda{=}1$ for \method{}-1st, $\alpha{=}\beta{=}1$ for \method{}-2nd, SNN latency $N{=}8$, hidden size 64, and learning rate $10^{-3}$. All experiments run on a single NVIDIA A6000 GPU.

\subsection{Performance Comparison}
\textbf{Homophilic Node Classification.}
Table~\ref{table_1} reports the homophilic results.
(1) Continuous-time methods (GraphCON and \method{}) outperform all discrete baselines, confirming the value of capturing subtle dynamics through continuous modelling.
(2) \method{}-1st and \method{}-2nd surpass most baselines: even with the information loss from spike encoding, our framework remains competitive on homophilic graphs while gaining substantial energy efficiency.
(3) \method{}-2nd consistently beats \method{}-1st, showing that high-order spike representation effectively mitigates the information-loss issue of first-order SNNs.
(4) \method{} also outperforms the pure-spike baseline (SpikingGCN) in most cases, indicating that adding the Graph ODE branch captures dynamic evolution without sacrificing the SNN energy advantage.

\begin{table}[t]
\centering
\caption{The test accuracy (in \%) for graph classification on MNIST datasets. \textbf{Bold} numbers indicate the best performance, and \underline{underline} indicates the second-best performance.}
\label{table_3}
\begin{tabular*}{0.65\linewidth}{@{\extracolsep{\fill}}lc}
\toprule
Model & Test accuracy \\
\midrule
ChebNet~\cite{defferrard2016convolutional} & 75.62 \\
MoNet~\cite{monti2017geometric} & 91.11 \\
PNCNN~\cite{finzi2022probabilistic} & 98.76 \\
SplineCNN~\cite{fey2018splinecnn} & 95.22 \\
GIN~\cite{xu2019powerful} & 97.23 \\
GatedGCN~\cite{bresson2017residual} & 97.95 \\
GCN~\cite{kipf2017semi} & 88.89 \\
GAT~\cite{velickovic2017graph} & 96.19 \\
GraphCON-GCN~\cite{rusch2022graph} & 98.68 \\
GraphCON-GAT~\cite{rusch2022graph} & $\underline{98.91}$ \\
\midrule
\method{}-1st & 98.82 \\
\method{}-2nd & $\textbf{98.92}$ \\
\bottomrule
\end{tabular*}
\end{table}

\textbf{Heterophilic Node Classification.}
Table~\ref{table_2} reports the heterophilic results.
(1) Message-passing baselines (GCN, GAT, GraphSAGE, Geom-GCN) trail the heterophily-aware methods (GPRGNN, H2GCN, GCNII, GraphCON, \method{}), as the neighbor-similarity assumption fails on these graphs and propagating heterophilic features hurts representation quality.
(2) \method{}-1st underperforms GraphCON because heterophilic prediction is more sensitive to feature precision, exposing the information loss of first-order SNNs.
(3) \method{}-2nd consistently outperforms \method{}-1st, again confirming the value of high-order spike representation in mitigating information loss.

\textbf{Graph Classification.}
Table~\ref{table_3} shows that continuous-time methods (\method{} and GraphCON) outperform all baselines, validating the benefit of modeling subtle dynamics on graph data. \method{}-2nd is best overall, while \method{}-1st sits between GraphCON-GCN and GraphCON-GAT: against GraphCON-GCN, the SNN binarization causes a small accuracy hit but yields large energy savings; against GraphCON-GAT, the absence of attention slightly limits \method{}-1st, but the high-order spike representation in \method{}-2nd recovers and exceeds it.

\section{Ablation Study}
\label{ablation_exp}
We further isolate the contribution of each high-order component by introducing two variants: \method{}-1st-2nd (first-order SNN + second-order Graph ODE) and \method{}-2nd-1st (second-order SNN + first-order Graph ODE), and report the full numbers in Appendix. Averaged across the six datasets, \method{}-2nd attains the best score (83.1), followed by \method{}-2nd-1st (80.5), \method{}-1st (80.1), and \method{}-1st-2nd (80.2). Two observations follow: (i) replacing the second-order SNN with a first-order one (\method{}-1st-2nd) is consistently the worst variant, confirming that mitigating spike-induced information loss is the dominant factor; (ii) \method{}-2nd-1st outperforms \method{}-1st and \method{}-1st-2nd, which suggests that high-order spike representation is more impactful than high-order Graph ODE for this task.

\subsection{Energy Efficiency Analysis}
\begin{figure}[ht]
\centering
\includegraphics[width=0.6\linewidth]{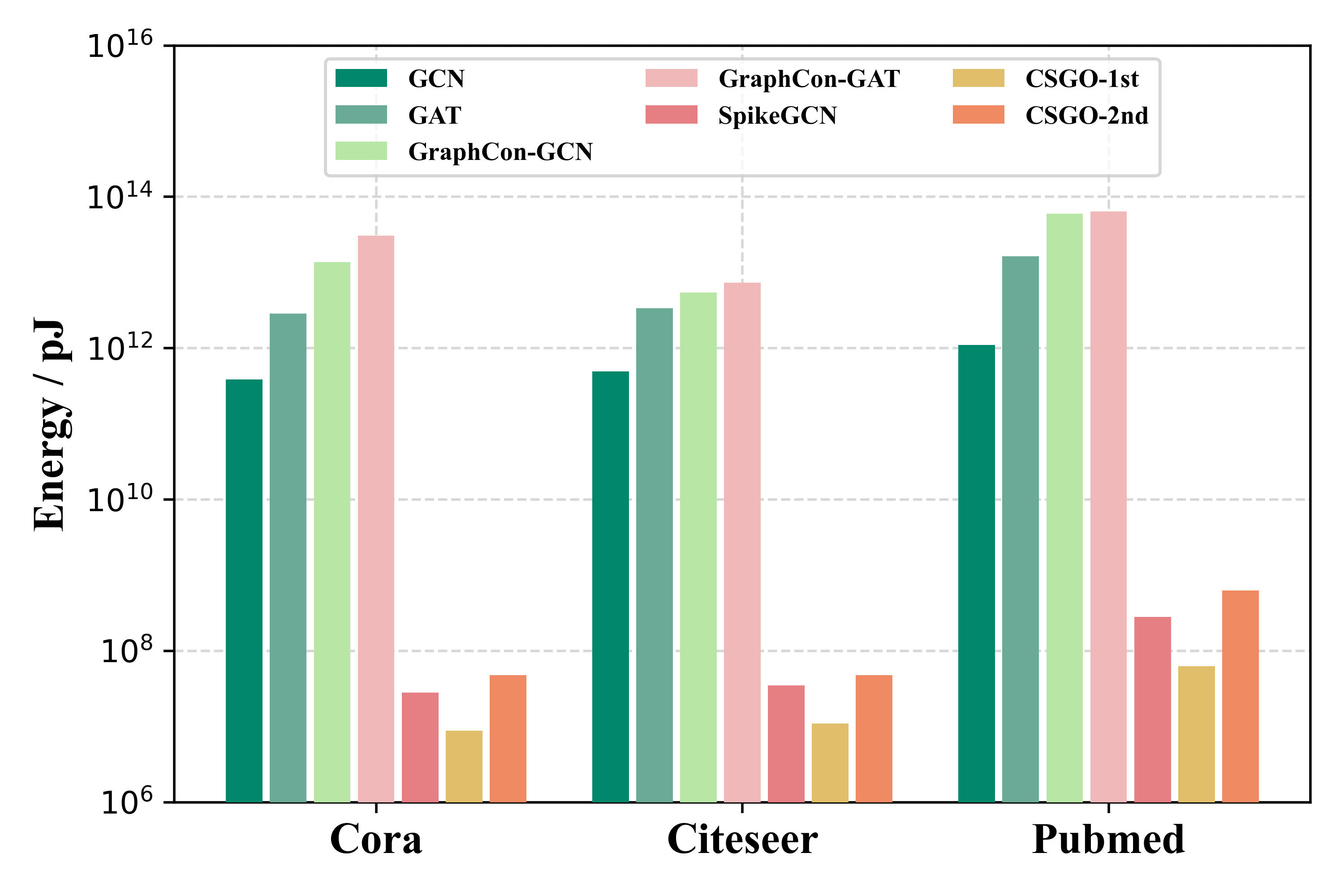}
\caption{Energy consumption comparison with various baselines on different datasets.}
\label{fig:energy_consumption}
\end{figure}

We measure energy consumption with the spike-counting metric of~\cite{zhu2022spiking,cao2015spiking} on three datasets: traditional GNNs are evaluated on an NVIDIA A6000 GPU, while spike-based models are evaluated on the ROLLS neuromorphic chip~\cite{indiveri2015neuromorphic}.
Figure~\ref{fig:energy_consumption} shows that (1) all spike-based methods (SpikeGCN, \method{}-1st, \method{}-2nd) consume significantly less energy than traditional GNNs, and (2) \method{}-1st is even more efficient than SpikeGCN, while \method{}-2nd costs slightly more but delivers substantially higher accuracy, making it the preferred deployment choice. Hyperparameter sensitivity is reported in Appendix.

\section{Conclusion}
We presented \textsc{CSGO}, a unified framework that couples Spiking Neural Networks with Graph ODEs for energy-efficient continuous-time graph learning. By bridging spike-latency dynamics with continuous graph evolution, \textsc{CSGO} enables expressive yet low-power representation learning, and we further introduced a second-order spiking variant that mitigates the information loss inherent in first-order SNNs. Theoretical analysis shows that gradients remain bounded throughout training, and experiments on homophilic, heterophilic, and graph-classification benchmarks confirm that \textsc{CSGO} matches or surpasses state-of-the-art accuracy while substantially reducing energy consumption. Future work will extend the framework to asynchronous event-based and multi-agent dynamic graphs.

\subsubsection{\discintname}
The authors have no competing interests to declare that are relevant to the
content of this article.

\section{Appendix}
\label{sec:appendix}

\paragraph{Proof of Proposition~1.}
The coupled first-order SNN and Graph ODE updates satisfy
\[
u_t^{\tau+1}
=
u_t^\tau
+
\int_\tau^{\tau+1}g(u_t^x,x)\,\mathrm{d}x,
\qquad
u_{t+1}^{\tau+1}
=
u_t^{\tau+1}
+
\int_t^{t+1}f(u_y^{\tau+1},y)\,\mathrm{d}y.
\]
Recursively expanding both dynamics and using the zero initial states gives
\[
u_T^N
=
2\int_0^{T-1}f(A_y)\,\mathrm{d}y
+
\int_{T-1}^{T}f(A_y)\,\mathrm{d}y,
\]
which proves Proposition~1.

\paragraph{Proof of Proposition~2.}
Unrolling the second-order SNN dynamics over the latency steps and
rearranging the synaptic convolution gives
\[
u(N)
=
\frac{
\sum_{n=0}^{N-1}
(\beta^{N-n}-\alpha^{N-n})I_{in}(n)
}{
\beta-\alpha
}
-
\sum_{n=0}^{N-1}
\beta^{N-n-1}V_{th}s(n).
\]
Using the spike-rate estimator with
$\alpha=e^{-\Delta\tau/\tau_{syn}}$ and
$\beta=e^{-\Delta\tau/\tau_{mem}}$, we obtain
\[
\hat a(N)
\approx
\operatorname{clamp}\left(
\frac{\tau_{syn}\tau_{mem}}
{\tau_{mem}-\tau_{syn}}
\hat I(N),
0,
\frac{V_{th}}{\Delta\tau}
\right).
\]
Following~\cite{meng2022training}, the residual membrane term is neglected.
Setting $\hat I(N)=\mathbf{Wz}$ yields Proposition~2, with
$\hat a^{i+1}(N)\approx\mathbf z^{i+1}$ as $N\to\infty$.

\paragraph{Proof of Proposition~3.}
By the definitions of $e(u_t^\tau)$ and $h(u_t^\tau)$,
\[
\frac{\partial u_t^\tau}{\partial\tau}=e(u_t^\tau),
\qquad
\frac{\partial u_t^\tau}{\partial t}=h(u_t^\tau).
\]
Integrating along both dimensions yields
\[
u_t^\tau
=
\int_0^T
h\left(
\int_0^N e(u_t^\tau)\,\mathrm{d}\tau
\right)\mathrm{d}t
=
\int_0^N
e\left(
\int_0^T h(u_t^\tau)\,\mathrm{d}t
\right)\mathrm{d}\tau,
\]
which proves Proposition~3.

\paragraph{Proof of Proposition~4.}
The Graph ODE gradient bound follows from~\cite{rusch2022graph}. For the
SNN parameters, the chain rule gives
\[
\frac{\partial\mathcal L}{\partial\mathbf W^k}
=
\frac{\partial\mathcal L}{\partial\mathbf Z_L^N}
\frac{\partial\mathbf Z_L^N}{\partial\mathbf Z_l^N}
\frac{\partial\mathbf Z_l^N}{\partial\mathbf Z_l^k}
\frac{\partial\mathbf Z_l^k}{\partial\mathbf W^k}.
\]
Using the corresponding Graph ODE and SNN bounds, we obtain
\[
\left|
\frac{\partial\mathcal L}{\partial\mathbf W^k}
\right|
\le
\frac{
(1+T\Gamma\Delta t)
(1+N\Theta\Delta\tau)V_{th}
}{
v\beta^2\Delta\tau
}A,
\]
which gives the second bound in Proposition~4.

\FloatBarrier
\subsection{Additional Experimental Results}

\paragraph{Full ablation results.}
Table~\ref{tab:appendix_ablation} reports the complete per-dataset results
for the ablation study discussed in the main paper.

\begin{table}[!ht]
\caption{Full ablation results. \textbf{Bold} numbers indicate the best performance.}
\label{tab:appendix_ablation}
\centering
\scriptsize
\setlength{\tabcolsep}{2.8pt}
\renewcommand{\arraystretch}{1.05}
\begin{tabular}{lccccccc}
\toprule
& Cora & Citeseer & Pubmed & Texas & Wisconsin & Cornell & Avg. \\
\textit{Homophily level}
& 0.81 & 0.74 & 0.80 & 0.11 & 0.21 & 0.30 & \\
\midrule
\method{}-1st-2nd
& 83.2$\pm$1.4 & 74.1$\pm$1.4 & 76.3$\pm$2.2
& 81.7$\pm$3.9 & 85.1$\pm$2.8 & 81.0$\pm$1.9 & 80.2 \\
\method{}-2nd-1st
& 83.5$\pm$1.8 & 73.4$\pm$2.1 & 77.2$\pm$2.3
& 83.1$\pm$3.8 & 84.4$\pm$2.2 & 81.2$\pm$2.7 & 80.5 \\
\method{}-1st
& 83.3$\pm$2.1 & 73.7$\pm$2.0 & 76.9$\pm$2.7
& 81.6$\pm$6.2 & 84.9$\pm$3.2 & 80.4$\pm$1.9 & 80.1 \\
\method{}-2nd
& \textbf{83.7$\pm$1.3} & \textbf{75.2$\pm$2.0}
& \textbf{79.6$\pm$2.3} & \textbf{87.3$\pm$4.2}
& \textbf{88.8$\pm$2.5} & \textbf{83.7$\pm$2.7}
& \textbf{83.1} \\
\bottomrule
\end{tabular}
\end{table}

\begin{figure}[!ht]
    \centering
    \includegraphics[width=0.8\textwidth]{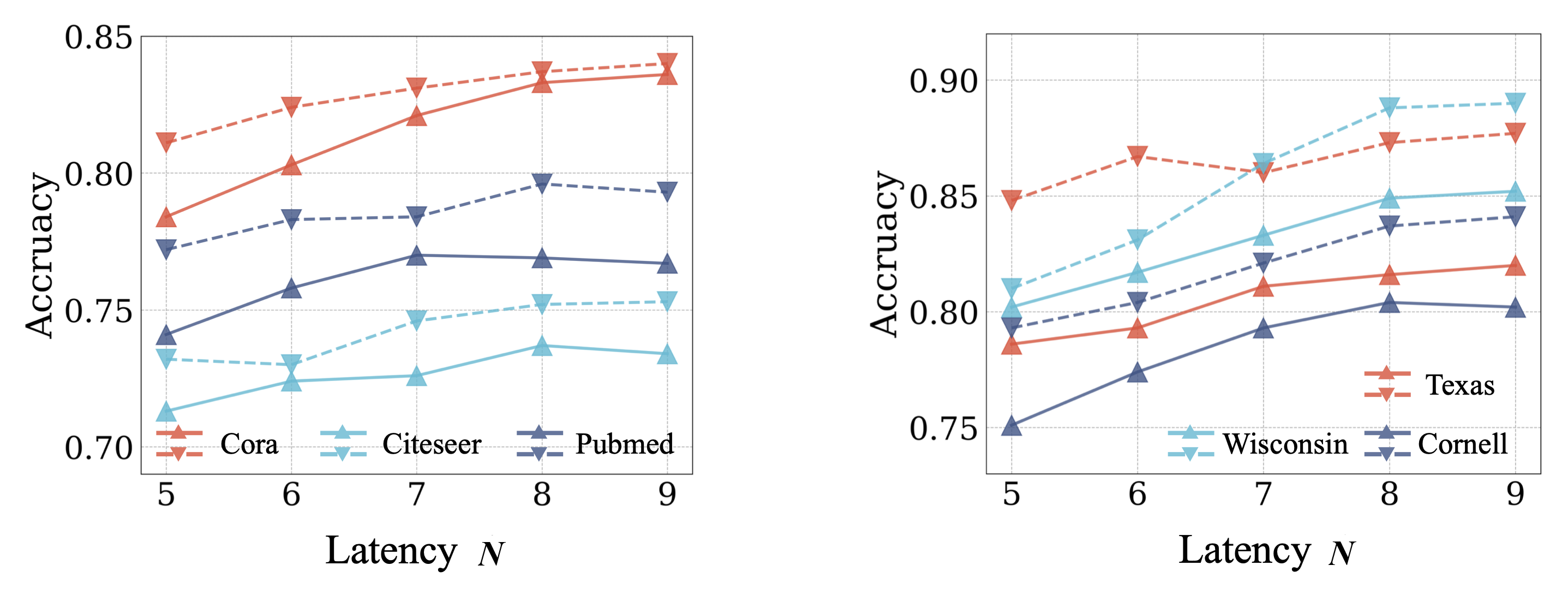}
    \caption{Sensitivity to SNN latency $N$. Solid and dotted lines denote
    \method{}-1st and \method{}-2nd, respectively.}
    \label{fig:appendix_latency}
\end{figure}

\paragraph{Sensitivity to SNN latency.}
We vary the SNN latency $N$ from 5 to 9 while keeping the other
hyperparameters fixed. As shown in Fig.~\ref{fig:appendix_latency},
performance generally improves and then stabilizes as $N$ increases.
Considering the trade-off between performance and computational cost,
we set $N=8$ by default.

\FloatBarrier

%
%
%
\bibliographystyle{splncs04}
\bibliography{example_paper}

@string{nips  = "Advances in Neural Information Processing Systems (NeurIPS)"}

@string{icml  = "International Conference on Machine Learning (ICML)"}

@string{iclr  = "International Conference on Learning Representations (ICLR)"}

@string{cvpr  = "IEEE Conference on Computer Vision and Pattern Recognition (CVPR)"}

@string{aaai  = "AAAI Conference on Artificial Intelligence"}

@string{ijcai = "International Joint Conference on Artificial Intelligence (IJCAI)"}

@string{kdd   = "ACM SIGKDD International Conference on Knowledge Discovery and Data Mining"}

@string{ijcv  = "International Journal of Computer Vision"}

@inproceedings{huang2021coupled,
  title={Coupled graph ode for learning interacting system dynamics},
  author={Huang, Zijie and Sun, Yizhou and Wang, Wei},
  booktitle=KDD,
  pages={705--715},
  year={2021}
}

@inproceedings{luo2023hope,
  title={HOPE: High-order graph ODE for modeling interacting dynamics},
  author={Luo, Xiao and Yuan, Jingyang and Huang, Zijie and Jiang, Huiyu and Qin, Yifang and Ju, Wei and Zhang, Ming and Sun, Yizhou},
  booktitle=ICML,
  pages={23124--23139},
  year={2023}
}

@inproceedings{kipf2017semi,
  title={Semi-supervised classification with graph convolutional networks},
  author={Kipf, Thomas N and Welling, Max},
  booktitle=ICLR,
  year={2017}
}

@inproceedings{xu2019powerful,
  title={How powerful are graph neural networks?},
  author={Xu, Keyulu and Hu, Weihua and Leskovec, Jure and Jegelka, Stefanie},
  booktitle=ICLR,
  year={2019}
}

@inproceedings{defferrard2016convolutional,
  title={Convolutional neural networks on graphs with fast localized spectral filtering},
  author={Defferrard, Micha{\"e}l and Bresson, Xavier and Vandergheynst, Pierre},
  booktitle=NIPS,
  year={2016}
}

@inproceedings{monti2017geometric,
  title={Geometric deep learning on graphs and manifolds using mixture model cnns},
  author={Monti, Federico and Boscaini, Davide and Masci, Jonathan and Rodola, Emanuele and Svoboda, Jan and Bronstein, Michael M},
  booktitle=CVPR,
  pages={5115--5124},
  year={2017}
}

@inproceedings{finzi2022probabilistic,
  title={Probabilistic numeric convolutional neural networks},
  author={Finzi, Marc Anton and Bondesan, Roberto and Welling, Max},
  booktitle=ICLR,
  year={2021}
}

@inproceedings{fey2018splinecnn,
  title={Splinecnn: Fast geometric deep learning with continuous b-spline kernels},
  author={Fey, Matthias and Lenssen, Jan Eric and Weichert, Frank and M{\"u}ller, Heinrich},
  booktitle=CVPR,
  pages={869--877},
  year={2018}
}

@article{bresson2017residual,
  title={Residual gated graph convnets},
  author={Bresson, Xavier and Laurent, Thomas},
  journal={arXiv preprint arXiv:1711.07553},
  year={2017}
}

@inproceedings{velickovic2017graph,
  title={Graph attention networks},
  author={Velickovic, Petar and Cucurull, Guillem and Casanova, Arantxa and Romero, Adriana and Lio, Pietro and Bengio, Yoshua and others},
  booktitle=ICLR,
  year={2017}
}

@inproceedings{rusch2022graph,
  title={Graph-coupled oscillator networks},
  author={Rusch, T Konstantin and Chamberlain, Ben and Rowbottom, James and Mishra, Siddhartha and Bronstein, Michael},
  booktitle=ICML,
pages={18888--18909},
  year={2022}
}

@article{pei2020geom,
  title={Geom-gcn: Geometric graph convolutional networks},
  author={Pei, Hongbin and Wei, Bingzhe and Chang, Kevin Chen-Chuan and Lei, Yu and Yang, Bo},
  journal={arXiv preprint arXiv:2002.05287},
  year={2020}
}

@article{mccallum2000automating,
  title={Automating the construction of internet portals with machine learning},
  author={McCallum, Andrew Kachites and Nigam, Kamal and Rennie, Jason and Seymore, Kristie},
  journal={Information Retrieval},
  volume={3},
  pages={127--163},
  year={2000},
  publisher={Springer}
}

@article{sen2008collective,
  title={Collective classification in network data},
  author={Sen, Prithviraj and Namata, Galileo and Bilgic, Mustafa and Getoor, Lise and Galligher, Brian and Eliassi-Rad, Tina},
  journal={AI magazine},
  volume={29},
  number={3},
  pages={93--93},
  year={2008}
}

@inproceedings{namata2012query,
  title={Query-driven active surveying for collective classification},
  author={Namata, Galileo and London, Ben and Getoor, Lise and Huang, Bert and Edu, U},
  booktitle={ International Workshop on Mining and Learning with Graphs},
  volume={8},
  pages={1},
  year={2012}
}

@article{lecun1998gradient,
  title={Gradient-based learning applied to document recognition},
  author={LeCun, Yann and Bottou, L{\'e}on and Bengio, Yoshua and Haffner, Patrick},
  journal={Proceedings of the IEEE},
  volume={86},
  number={11},
  pages={2278--2324},
  year={1998},
  publisher={Ieee}
}

@inproceedings{xhonneux2020continuous,
  title={Continuous graph neural networks},
  author={Xhonneux, Louis-Pascal and Qu, Meng and Tang, Jian},
  booktitle=ICML,
  year={2020}
}

@inproceedings{chen2018neural,
  title={Neural ordinary differential equations},
  author={Chen, Ricky TQ and Rubanova, Yulia and Bettencourt, Jesse and Duvenaud, David K},
  booktitle=NIPS,
  year={2018}
}

@article{poli2019graph,
  title={Graph neural ordinary differential equations},
  author={Poli, Michael and Massaroli, Stefano and Park, Junyoung and Yamashita, Atsushi and Asama, Hajime and Park, Jinkyoo},
  journal={arXiv preprint arXiv:1911.07532},
  year={2019}
}

@inproceedings{zhu2020generalizing,
  title={Generalizing graph neural networks beyond homophily},
  author={Zhu, Jiong and Yan, Yujun and Zhao, Lingxiao and Heimann, Mark and Akoglu, Leman and Koutra, Danai},
  booktitle=NIPS,
  year={2020}
}

@article{eshraghian2023training,
  title={Training spiking neural networks using lessons from deep learning},
  author={Eshraghian, Jason K and Ward, Max and Neftci, Emre O and Wang, Xinxin and Lenz, Gregor and Dwivedi, Girish and Bennamoun, Mohammed and Jeong, Doo Seok and Lu, Wei D},
  journal={Proceedings of the IEEE},
  year={2023},
  publisher={IEEE}
}

@inproceedings{meng2022training,
  title={Training high-performance low-latency spiking neural networks by differentiation on spike representation},
  author={Meng, Qingyan and Xiao, Mingqing and Yan, Shen and Wang, Yisen and Lin, Zhouchen and Luo, Zhi-Quan},
  booktitle=CVPR,
  pages={12444--12453},
  year={2022}
}

@inproceedings{kipf2018neural,
  title={Neural relational inference for interacting systems},
  author={Kipf, Thomas and Fetaya, Ethan and Wang, Kuan-Chieh and Welling, Max and Zemel, Richard},
  booktitle=ICML,
  pages={2688--2697},
  year={2018}
}

@inproceedings{xu2021exploiting,
  title={Exploiting spiking dynamics with spatial-temporal feature normalization in graph learning},
  author={Xu, Mingkun and Wu, Yujie and Deng, Lei and Liu, Faqiang and Li, Guoqi and Pei, Jing},
  booktitle = IJCAI,
pages        = {3207--3213},
  year      = {2021}
}

@inproceedings{zhu2022spiking,
      title={Spiking Graph Convolutional Networks}, 
      author={Zulun Zhu and Jiaying Peng and Jintang Li and Liang Chen and Qi Yu and Siqiang Luo},
      booktitle=IJCAI,
pages        = {2434--2440},
      year={2022}
}

@inproceedings{li2023scaling,
  title={Scaling Up Dynamic Graph Representation Learning via Spiking Neural Networks},
  author={Li, Jintang and Yu, Zhouxin and Zhu, Zulun and Chen, Liang and Yu, Qi and Zheng, Zibin and Tian, Sheng and Wu, Ruofan and Meng, Changhua},
  booktitle=AAAI,
pages        = {8588--8596},
  year={2023}
}

@inproceedings{yin2024dynamic,
  title={Dynamic Spiking Graph Neural Networks},
  author={Yin, Nan and Wang, Mengzhu and Chen, Zhenghan and De Masi, Giulia and Xiong, Huan and Gu, Bin},
  booktitle=AAAI,
  pages={16495--16503},
  year={2024}
}

@inproceedings{yan2021near,
  title={Near lossless transfer learning for spiking neural networks},
  author={Yan, Zhanglu and Zhou, Jun and Wong, Weng-Fai},
  booktitle=AAAI,
  pages        = {10577--10584},
  year={2021}
}

@inproceedings{indiveri2015neuromorphic,
  title={Neuromorphic architectures for spiking deep neural networks},
  author={Indiveri, Giacomo and Corradi, Federico and Qiao, Ning},
  booktitle={IEEE International Electron Devices Meeting (IEDM)},
  pages={4--2},
  year={2015}
}

@article{cao2015spiking,
  title={Spiking deep convolutional neural networks for energy-efficient object recognition},
  author={Cao, Yongqiang and Chen, Yang and Khosla, Deepak},
  journal=IJCV,
  volume={113},
  pages={54--66},
  year={2015}
}

\end{document}